\documentclass[runningheads]{llncs}
\usepackage[T1]{fontenc}
\usepackage{graphicx}
\usepackage{hyperref}
\usepackage{color}

\usepackage{multirow} 
\usepackage{subcaption} 
\usepackage{booktabs} 
\usepackage[table]{xcolor} 
\usepackage{hhline}

\usepackage{amssymb} 
\usepackage{mathtools} 

\DeclareMathOperator{\agg}{agg}
\DeclareMathOperator*{\argmin}{\arg\!\min} 

\DeclarePairedDelimiterX{\infdivx}[2]{(}{)}{%
  #1\;\delimsize\|\;#2%
}

\usepackage[size=footnotesize]{todonotes}

\begin{document}
\title{Trusting Fair Data: Leveraging Quality in Fairness-Driven Data Removal Techniques\thanks{Manuscript submitted to the 26th International Conference on Big Data Analytics and Knowledge Discovery (DaWaK 2024).}}
\titlerunning{Trusting Fair Data}
%
\author{Manh Khoi Duong\inst{1}\orcidID{0000-0002-4653-7685} \and \\
Stefan Conrad\inst{1}\orcidID{0000-0003-2788-3854}}
\authorrunning{M. K. Duong \and S. Conrad}
\institute{Heinrich Heine University, Universit\"atsstra\ss{}e 1, 40225 D\"usseldorf, Germany
\email{\{manh.khoi.duong, stefan.conrad\}@hhu.de}}

%
%
%
\maketitle              
\begin{abstract}
    In this paper, we deal with bias mitigation techniques that
    remove specific data points from the training set to aim for
    a fair representation of the population in that set.
    Machine learning models are trained on these pre-processed datasets, and their predictions are expected to be fair.
    However, such approaches may exclude relevant data, making the attained subsets
    less trustworthy for further usage.
    To enhance the trustworthiness of prior methods, we propose additional requirements and objectives that the subsets must fulfill in addition
    to fairness: (1) group coverage, and (2) minimal data loss.
    While removing entire groups may improve the measured fairness,
    this practice is very problematic as failing to represent every group
    cannot be considered fair.
    In our second concern, we advocate for the retention of data while minimizing discrimination.
    By introducing a multi-objective optimization problem that considers fairness and data loss, we propose a methodology to find Pareto-optimal solutions that balance these objectives.
    By identifying such solutions,
    users can make informed decisions about the trade-off between fairness and data quality and select the most suitable subset for their application.
    Our method is distributed as a Python package via PyPI 
    under the name \texttt{FairDo}\footnote{https://github.com/mkduong-ai/fairdo}.

\end{abstract}
\section{Introduction}
Machine learning models are often trained on biased data, which can lead to biased predictions~\cite{catania2023fairness}.
A common approach to addressing fairness concerns is to use bias mitigation techniques.
They can be categorized into pre-processing, in-processing, and post-processing~\cite{mehrabi2021survey}.
Pre-processing techniques aim to remove bias from the training data
before training a machine learning model, in-processing techniques
modify the learning algorithm, and post-processing techniques adjust
the predictions.

Because bias can be introduced at various stages during data preparation steps, pre-processing techniques can be integrated into the data preparation pipeline to ensure that the training data is fair.
Some of the pre-processing techniques involve the removal of
certain data points from the training set~\cite{verma2021removing,duong2023gen,duong2023framework}.
By removing certain data points, the machine learning model is trained on a fair subset and its predictions are expected to be fair as well.
These techniques aim to fix representation bias in the data.
While they tackle the root cause of the problem, they 
are deemed problematic as they may lead to the exclusion of relevant data.

In this paper, we explore multiple problems that can arise from the
removal of data points and propose a decision-making methodology
for selecting a subset that is more trustworthy for the user.
One particular problem is the \emph{removal of groups} as a whole,
i.e., lack of \emph{coverage}.
For instance, the elimination of non-privileged groups
can be considered fair by various fairness metrics,
yet it fails to align with our main objective, which is to
represent every group fairly in the resulting dataset.
Removing entire groups can be seen as equivalent
to underreporting discrimination.
Another problem we tackle is the \emph{amount of data removed}.
When removing too many data points, the resulting dataset
may not accurately represent the original data
and data quality is compromised.
Hence, our second objective is to retain as much data
as possible.
With this additional objective, there is a trade-off between
fairness and the amount of data removed.

In summary, our main contributions are as follows:
\begin{itemize}
    \item We present two additional criteria,
    \emph{coverage} and \emph{data loss},
    to enhance trustworthiness for fairness-driven data removal techniques.
    \item We propose a multi-objective optimization problem that considers fairness and data loss. Using NSGA-II~\cite{deb2002nsga2},
    we find Pareto-optimal solutions that the user can choose from.
    \item We provide an extensive and empirical evaluation of
    our proposed methodology on three real-world datasets
    (Adult~\cite{ron1996_adult}, Bank~\cite{moro2014bank},
    COMPAS~\cite{larson_angwin_mattu_kirchner_2016}) and evaluate
    the attained subsets by training machine learning models on them.
    We assess the models' fairness and performances by
    comparing them to models trained on the original datasets.
    \item We publish our methods in an open-source and documented
    Python package \texttt{FairDo}
    that can be used on-the-fly to pre-process datasets.
    It comes with several tutorials and examples.
\end{itemize}

\section{Related Work}
While there are many bias mitigation techniques, and some of them
are implemented in popular Python packages such as \texttt{AIF360}~\cite{aif360}
and \texttt{Fairlearn}~\cite{fairlearn},
the included pre-processing
techniques are not able to deal with \emph{non-binary groups}
and at the same time transform the data in an \emph{uninterpretable}
way by editing features and labels.
The package \texttt{FairDo}~\cite{duong2023framework} aims to address
these issues.
It is a highly adaptive framework for removing data points to achieve fairness.
In this framework, it is possible to deal with
binary, non-binary groups, and multiple protected attributes if
the fairness metric is defined accordingly.
While the resulting fair datasets are more interpretable, as they
are subsets of the original dataset, data removal can also be
viewed critically.
One can argue that their framework offers other solutions,
such as adding synthetic data points to the original data for fairness.
However, all solutions are based on data removal within their framework.

Drawing inspiration from prior research that has addressed group
\emph{representativeness}~\cite{drosou2017diversity,stoyanovich2018online,catania2023represent},
we extend the methodology of Duong et al.~\cite{duong2023framework}
accordingly.
There are many ways to define representativeness.
The work of Stoyanovich et al.~\cite{stoyanovich2018online} explores
two main definitions: \emph{proportional
representation} and \emph{coverage}.
Proportional representation ensures that the dataset
contains a representative number of data points from each group.
Coverage, on the other hand, ensures that the dataset
covers the entire population.
We aim for coverage in our work and argue
that aiming for proportional representation
is not feasible because the group counts are given
beforehand and not much can be done about it during the
data removal process.
Coverage only requires each group to be represented at least
once in the dataset and is hence a more relaxed constraint.

The work of Catania et al.~\cite{catania2023represent}
is also related to our work.
They propose a constraint-based optimization approach
for their problem, which is to mitigate biases in datasets
during a selection-based query.
We deal with selecting a subset
that optimizes a certain fairness objective and do not
consider being in a query setting.
\section{Preliminaries}\label{section:preliminaries}
Following definitions are primarily based on the work of
Žliobaitė~\cite{liobait2017MeasuringDI} and taken from
Duong et al.~\cite{duong2023framework}.
We applied minor modifications to fit the definitions to the
context of this paper.

\subsection{Measuring Discrimination}
Protected attributes such as race, gender, and nationality make individuals
vulnerable to discrimination.
Generally, we use $Z$ to represent a protected attribute
and $Y$ to denote the outcome for an individual.
Formally, we define $Z$ and $Y$ as discrete random variables.
$Z$ can take on values from the sample space $g$, which represents
social groups such as male, female, and non-binary.
For $Y$, we use the values 1 and 0 to indicate positive and negative outcomes, respectively.
Further, we denote $z_i$ and $y_i$ to refer to the values
of the $i$-th individual.

\begin{definition}[Dataset]
    We define a dataset $\mathcal{D}$ as a set of data points $d_i$:
    \begin{equation*}
        \mathcal{D} = \{d_i\}_{i=1}^n,
    \end{equation*}
    where $n$ is the number of data points in the dataset.
    A data point can be defined as a triplet $(x_i, z_i, y_i)$, where $x_i$ is the feature vector, $z_i$ is the protected attribute, and $y_i$ is the outcome.
\end{definition}
Fairness criteria are often based on conditional probabilities,
and typically demand some equal outcome between groups~\cite{hardt2016equality,zafar2017-disparate}.
One of the most common fairness criteria is \emph{statistical parity}~\cite{calders2009}.
\begin{definition}[Statistical Parity~\cite{calders2009}]
    Statistical parity
    requires equal positive outcomes between groups:
    \begin{equation*}
        P(Y=1 \mid Z=i) = P(Y=1 \mid Z=j),
    \end{equation*}
    where $i, j \in g$ represent different groups.
\end{definition}
Typically, the probabilities are estimated using sample statistics.
Because achieving equal probabilities for
certain outcomes is not always possible,
existing literature~\cite{liobait2017MeasuringDI}
present measures to quantify the level of discrimination.
\begin{definition}[Statistical Disparity~\cite{liobait2017MeasuringDI}]\label{eq:statdis}
    \emph{Statistical disparity} is defined as the absolute difference between the probabilities of the positive outcome $Y=1$ between two groups $i, j \in g$:
    \begin{equation*}
        \delta_{Z}(i, j) = |P(Y=1 \mid Z=i) - P(Y=1 \mid Z=j)|.
    \end{equation*}
\end{definition}
Establishing $\delta_{Z}$ provides a fundamental foundation for various scenarios.
For instance, it allows us to aggregate pairwise differences between groups,
particularly when dealing with attributes that are non-binary~\cite{liobait2017MeasuringDI}.
This allows us to quantify discrimination for more than two groups.
\begin{definition}[Disparity for Non-binary Groups]\label{def:aggnonbinary}
    We introduce an aggregate function $\agg^{(1)}$ as a function that takes a set of values and returns a single value.
    The function $\agg^{(1)}$ can represent, for example, the sum or maximum function.
    With $\agg^{(1)}$, we can compute the discrimination for a single protected attribute $Z$ with any amount of groups.
    Simplifying notation, we write $\psi(\mathcal{D})$ to represent the discrimination measure for a dataset $\mathcal{D}$:
    \begin{equation*}
	    \psi(\mathcal{D}) = \underset{i, j \in g, i \neq j}{\agg^{(1)}} \delta_{Z}(i, j).
    \end{equation*}
\end{definition}

\begin{example}[Maximal Statistical Disparity]\label{ex:maxstatdisparity}
    The maximal statistical disparity is defined as:
    \begin{equation*}
        \psi_\text{SDP-max}(\mathcal{D}) = \max_{i, j \in g, i \neq j} \delta_{Z}(i, j).
    \end{equation*}
    It describes the maximum discrimination obtainable between two groups.
\end{example}
There are many ways to measure discrimination in a dataset.
In this paper, we focus on the maximal statistical disparity as it provides
an interpretable measure of discrimination and is recommended
in the work of Žliobaitė~\cite{liobait2017MeasuringDI}.
Still, our framework can be extended to other measures as well,
that is, any $\psi$ that maps a dataset to a positive value.
Our framework only assumes that the objective is to minimize $\psi$.
We use the term \emph{discrimination measure} and \emph{fairness metric}
interchangeably to refer to $\psi$.



\subsection{Fair Subset Selection}
Duong et al.~\cite{duong2023framework} proposed a framework for removing
discriminating data points from a given dataset $\mathcal{D} = \{d_i\}_{i=1}^n$.
They stated the problem as finding a subset
$\mathcal{D}_\text{fair} \subseteq \mathcal{D}$, which minimizes
the discrimination in that subset.
This describes following combinatorial optimization problem:
\begin{equation}
    \min_{\mathcal{D}_\text{fair} \subseteq \mathcal{D}} \quad \psi(\mathcal{D}_\text{fair}).
\end{equation}
To make the problem solvable,
the authors~\cite{duong2023framework} introduced a binary
decision variable $b_i$ for each sample
$d_i \in \mathcal{D}$, where $b_i = 1$ indicates if the sample
$d_i$ is included in the fair subset $\mathcal{D}_\text{fair}$
and 0 otherwise. More formally $\mathcal{D}_\text{fair}$ is defined as:
\begin{equation}
    \mathcal{D}_\text{fair} = \{d_i \in \mathcal{D} \mid b_i = 1\}.
\end{equation}
Defining $\mathbf{b} = (b_1, b_2, \ldots, b_n) \in \{0, 1\}^n$ as a solution vector, finding the optimal subset $\mathcal{D}_\text{fair}$ is equivalent to solving for the optimal binary vector $\mathbf{b}^*$:
\begin{equation}
    \mathbf{b}^* = \argmin_{\mathbf{b} \in \{0, 1\}^n} \quad \psi(\{d_i \in \mathcal{D} \mid b_i = 1\})
\end{equation}

Because finding the exact optimal solution $\mathbf{b}^*$ is an
NP-hard problem if $\psi$ is treated as a black-box,
the authors~\cite{duong2023framework}
employed genetic algorithms to heuristically solve the problem.
\section{Enhancing Trust in Fair Data}
When having a fair dataset, we want to ensure that the dataset is still
faithful to its original version.
To enhance the trustworthiness,
we introduce two additional criteria.
The two criteria roughly describe a form of quality assurance
for the fair subset.
Overall, we have three criteria:
\begin{itemize}
    \item \textbf{Fairness}: We want to minimize the discrimination in the attained subset~\cite{duong2023framework}.
    \item \textbf{Coverage}: All groups
    must be included in the fair subset.
    \item \textbf{Data Loss}: The fair subset
    should resemble the original dataset by retaining
    as much data as possible.
\end{itemize}

\subsection{Coverage}\label{section:coverage}
When we compare the discrimination scores between two datasets,
one could naively assume that the dataset with the lower
score should be preferred over the other.
However, simply comparing the discrimination scores is not sufficient.

\subsubsection{Example}
Table~\ref{table:original} represents the original dataset $\mathcal{D}$, and Table~\ref{table:fairsubset}
depicts a subset $\mathcal{D}_\text{fair} \subseteq \mathcal{D}$,
purposely selected to achieve fairness.
The discrimination scores are $\psi_\text{SDP-max}(\mathcal{D}) = 0.5$ and $\psi_\text{SDP-max}(\mathcal{D}_\text{fair}) = 0$.
Despite $\mathcal{D}_\text{fair}$ yielding a perfect fairness score,
group 2 is missing in that set.
Another fair subset $\mathcal{D}_\text{cov.} \subseteq \mathcal{D}$
is shown in Table~\ref{table:fairsubsetcoverage},
which includes all groups and hence satisfies coverage.
It also achieves a perfect fairness score, $\psi_\text{SDP-max}(\mathcal{D}_\text{cov.}) = 0$.
We even argue that any dataset $\mathcal{D}_\text{cov.}$ satisfying coverage is more preferred than any other dataset
that does not, regardless of their fairness scores.

\begin{table}[tb]
    \caption{A dataset and two of its possible subsets $\mathcal{D}_\text{fair}, \mathcal{D}_\text{cov.} \subseteq \mathcal{D}$. Both subsets achieve perfect fairness scores but only $\mathcal{D}_\text{cov.}$ satisfies coverage.}
    \begin{minipage}{.32\linewidth}
      \centering
      \caption{$\mathcal{D}$}
      \begin{tabular}{|c|c|c|}
    	\hline
    	\textbf{$d_i$} & \textbf{z} & \textbf{y} \\
	    \hline
	    1 & 0 & 1 \\
	    2 & 1 & 1 \\
	    3 & 2 & 0 \\
	    4 & 2 & 1 \\
    	\hline
	  \end{tabular}
      \label{table:original}
    \end{minipage}%
    \begin{minipage}{.32\linewidth}
      \centering
      \caption{$\mathcal{D}_\text{fair}$}
      \begin{tabular}{|c|c|c|}
        \hline
        \textbf{$d_i$} & \textbf{z} & \textbf{y} \\
        \hline
        1 & 0 & 1 \\
        2 & 1 & 1 \\
        \rowcolor{gray!25}\textcolor{gray!50}{3} & \textcolor{gray!50}{2} & \textcolor{gray!50}{0} \\
	    \rowcolor{gray!25}\textcolor{gray!50}{4} & \textcolor{gray!50}{2} & \textcolor{gray!50}{1} \\
        \hline
      \end{tabular}
      \label{table:fairsubset}
    \end{minipage}
    \begin{minipage}{.32\linewidth}
        \centering
        \caption{$\mathcal{D}_\text{cov.}$}
        \begin{tabular}{|c|c|c|}
            \hline
            \textbf{$d_i$} & \textbf{z} & \textbf{y} \\
            \hline
            1 & 0 & 1 \\
            2 & 1 & 1 \\
            \rowcolor{gray!25}\textcolor{gray!50}{3} & \textcolor{gray!50}{2} & \textcolor{gray!50}{0} \\
            4 & 2 & 1 \\
            \hline
        \end{tabular}
        \label{table:fairsubsetcoverage}
    \end{minipage}
\end{table}

\subsubsection{Incorporating Coverage}
We want to construct a fairness metric $\hat{\psi}$ that reflects
our preference regarding coverage: The subset that satisfies coverage
is always preferred over the subset that does not.
However, if both subsets satisfy coverage, we want to compare them based on the fairness metric $\psi$.
Thus, a penalty is only applied if a group is missing.
In this case, $\hat{\psi}$ must have a higher value
than the maximum discrimination achievable in $\psi$
to enforce the preference.

\begin{definition}[Penalized Discrimination]\label{def:penalty-max}
The highest disparity possible for $\psi_\text{SDP-max}$ is $1$.
Let $|g_m|$ be the number of missing groups and $\epsilon > 0$, then
we penalize $\psi_\text{SDP-max}$ as follows to enforce
preferring coverage over non-coverage:
\begin{equation*}
    \hat{\psi}_\text{SDP-max}(\mathcal{D}) = \max(\psi_\text{SDP-max}(\mathcal{D}), [|g_m| > 0] \cdot (1+\epsilon)),
\end{equation*}
where $[|g_m| > 0]$ is an indicator function, which returns $1$ if $|g_m| > 0$ and $0$ otherwise.
Setting $\epsilon$ to any positive value ensures that the penalty is higher than the maximum discrimination score.
\end{definition}

\subsection{Data Loss}
By data loss, we refer to the similarity between the fair subset
and its original.
There are several ways to measure data loss by this means.
But some methods require knowledge of the true underlying
distribution of the data, which has to be expensively estimated.
Therefore, we use an efficient measure, which is the
relative amount of data removed.

\begin{definition}[Data Loss]\label{def:dataloss}
The relative amount of the data removed is given by:
\begin{equation*}
    \mathcal{L}(\mathcal{D}, \mathcal{D}_\text{fair}) = 1 - \frac{|\mathcal{D}_\text{fair}|}{|\mathcal{D}|},
\end{equation*}
where $\mathcal{D}_\text{fair} \subseteq \mathcal{D}$.
A lower value indicates less data is removed, and therefore is better.
\end{definition}

\section{Optimization Objectives}
We have three objectives to optimize for: \emph{fairness}, \emph{coverage},
and \emph{data loss}.
Two of the objectives, fairness and coverage, can be combined into a single objective, as shown in Definition~\ref{def:penalty-max}.
The third objective, data loss, can be treated in multiple ways,
which we will discuss in the following.

\subsection{Multi-objective Optimization}
If there are no preferences provided regarding fairness and data loss,
we have to treat the problem as a multi-objective optimization problem.
The aim is to minimize both discrimination $\hat{\psi}$ and data loss $\mathcal{L}$.
The optimization problem is written as follows:
\begin{equation}\label{eq:multiobjective}
    \min_{\mathcal{D}_\text{fair} \subseteq \mathcal{D}} \quad
        (\hat{\psi}(\mathcal{D}_\text{fair}), \mathcal{L}(\mathcal{D},\mathcal{D}_\text{fair})).
\end{equation}
Solvers for multi-objective optimization problems aim to find the Pareto front, which is the set of solutions that are not dominated
by any other solution. A solution is dominated if there is another solution that is better in at least one objective and not worse in any other objective.

\subsection{Single-objective Optimization}
If the importance of fairness and data loss is known beforehand,
the problem can be transformed into a single-objective optimization problem by using the weighted sum of the objectives:
\begin{equation}
    \min_{\mathcal{D}_\text{fair} \subseteq \mathcal{D}} \quad \alpha \hat{\psi}(\mathcal{D}_\text{fair}) + (1-\alpha) \mathcal{L}(\mathcal{D},\mathcal{D}_\text{fair}),
\end{equation}
where $\alpha \in [0, 1]$ is a weighting factor that determines the importance of fairness over data loss.
A value of $\alpha = 0.5$ indicates that both objectives are equally important and is set as the default value in our experiments.
When $\alpha$ values are set lower, the user prioritizes data fidelity more than fairness.

However, both objectives do not necessarily map to the same scale
and therefore require normalization to make them comparable.
Specifically, $\hat{\psi}$ requires normalization depending
on the fairness metric used. Introducing $\beta$ as the
normalization factor, the single-objective optimization problem is then:
\begin{equation}\label{eq:singleobjective}
    \min_{\mathcal{D}_\text{fair} \subseteq \mathcal{D}} \quad \frac{\alpha \hat{\psi}(\mathcal{D}_\text{fair})}{\beta} + (1-\alpha) \mathcal{L}(\mathcal{D},\mathcal{D}_\text{fair}).
\end{equation}
There are two meaningful choices for $\beta$: We either care about
the \emph{absolute} or \emph{relative discrimination} score compared to the original dataset.
For the absolute score, $\beta$ is set as the theoretical maximum value of $\hat{\psi}$.
In the case of $\hat{\psi}_\text{SDP-max}$, setting $\beta = 1$ or $\beta = 1 + \epsilon$ are both viable and similar options if
$\epsilon$ is small enough.
For the relative score, $\beta$ is set as the discrimination score of the original dataset, $\beta = \hat{\psi}(\mathcal{D})$.
Hence, any discrimination score under 1 implies a reduction of discrimination. The score can be interpreted as the percentage of discrimination removed or added.

We note that normalizing is not required in the multi-objective optimization approach as the objectives are treated separately
and the used heuristic compares candidate solutions
based on the Pareto order.
Only selecting a single solution from the Pareto front
requires weighting the objectives.

\section{Heuristics}\label{section:ga}
To make the framework flexible and agnostic to the fairness metric,
we have to use heuristics that only require function evaluations.
For this, we use genetic algorithms to solve the optimization problems
in Equations~\eqref{eq:multiobjective} and~\eqref{eq:singleobjective}.
To solve the single-objective problem, we use the genetic algorithm
borrowed from Duong et al.~\cite{duong2023framework}.
For the multi-objective optimization problem, we use our own modified
version of the NSGA-II~\cite{deb2002nsga2} algorithm as described in the following.

\subsection{NSGA-II Modification}
We added the possibility to
select between methods that initialize the population in our modified
NSGA-II~\cite{deb2002nsga2} algorithm.
We created our own initializer that initializes the population with variable bias.

Generally, all employed algorithms operate on a population of solutions.
In our implementation, the population is encoded as a binary matrix $\mathbf{P} \in \{0, 1\}^{M \times n}$,
where each row $\mathbf{b}_i \in \{0, 1\}^n$ represents a solution, i.e.,
\begin{equation}
    \mathbf{P} = ( \mathbf{b}_1, \mathbf{b}_2, \ldots, \mathbf{b}_M)^\top,
\end{equation}
and $M$ is the population size.

\subsubsection{Random Initializer}
A common approach to initialize a population is to randomly assign each entry in the
binary matrix $\mathbf{P}$ to 0 or 1 with a certain probability $p$~\cite{introductionevolution}.
Usually, the probability is set to 0.5.
Trivially, for all $\mathbf{b}_i \in \mathbf{P}$, the expected number of 1s is:
\begin{equation}
    \mathbb{E}(\mathbf{b}_i) \rightarrow \frac{n}{2},
\end{equation}
as $n$ approaches infinity.
This implies that each subset is expected to be half the size of the original dataset.

Hence, this initialization method is not suitable for our problem, as we aim to minimize the number of removed data points and require a diverse population.

\subsubsection{Variable Initializer}
To address the issues with the prior method,
we propose an initializer, which creates individuals with
varying probabilities $p$.
The user can specify the range of the probabilities $[p_\text{min}, p_\text{max}]$,
and the initializer
will create individuals with probabilities evenly distributed within the specified range.
This leads to a more diverse population, where some individuals have less
data removed than others.
The default range is set to $[p_\text{min}, p_\text{max}] = [0.5, 0.99]$.





\section{Evaluation}\label{sec:evaluation}
To evaluate the proposed framework, we conducted multiple experiments.
Following research questions guided our evaluation:
\begin{itemize}
    \item \textbf{RQ1} Which configuration of genetic operators is best suited for the NSGA-II algorithm in the context of bias mitigation in datasets?
    \item \textbf{RQ2} What is the impact of pre-processed datasets
    on the fairness and performance of machine learning models,
    as compared to models trained on unprocessed data?
\end{itemize}
Each following subsection corresponds to one of the research questions.
Each experiment was conducted on the same objectives, datasets,
and other settings as listed below.
The specific details for each experiment are detailed in the
corresponding subsections.

\subsubsection{Objectives}
The objectives are
$\hat{\psi}_\text{SDP-max}$ (Definition~\ref{def:penalty-max})
with $\epsilon = 0.01$ and
$\mathcal{L}$ (Definition~\ref{def:dataloss}).
Both are to be minimized.

\subsubsection{Datasets}
We conducted all experiments on three popular datasets in the fairness literature:
Adult~\cite{ron1996_adult}, Bank~\cite{moro2014bank},
and Compas~\cite{larson_angwin_mattu_kirchner_2016}, providing a comprehensive examination across various domains.
They all serve as baselines for our experiments.

\subsubsection{Trials}
We conducted 10 trials for each configuration in the experiments
to ensure the reliability of our results.


\begin{table}[tb]
    \caption{Hyperparameter optimization results showing
    hypervolume indicator values for different genetic operators.
    Best results are highlighted in bold.}
    \label{table:results_hyper}
    \begin{tabular}{llll|lll}
    \toprule
    Initializer & Selection & Crossover & Mutation & Adult & Bank & Compas \\
    \midrule
    \multirow[t]{8}{*}{Random} & \multirow[t]{4}{*}{Elitist} & \multirow[t]{2}{*}{1-Point} & Bit Flip & 0.49 ± 0.00 & 0.47 ± 0.00 & 0.53 ± 0.00 \\
     &  &  & Shuffle & 0.47 ± 0.01 & 0.44 ± 0.01 & 0.52 ± 0.03 \\
    \cline{3-7}
     &  & \multirow[t]{2}{*}{Uniform} & Bit Flip & 0.48 ± 0.00 & 0.45 ± 0.00 & 0.51 ± 0.01 \\
     &  &  & Shuffle & 0.46 ± 0.01 & 0.43 ± 0.01 & 0.49 ± 0.01 \\
    \cline{2-7} 
     & \multirow[t]{4}{*}{Tournament} & \multirow[t]{2}{*}{1-Point} & Bit Flip & 0.49 ± 0.00 & 0.47 ± 0.00 & 0.53 ± 0.00 \\
     &  &  & Shuffle & 0.50 ± 0.01 & 0.46 ± 0.00 & 0.57 ± 0.02 \\
    \cline{3-7}
     &  & \multirow[t]{2}{*}{Uniform} & Bit Flip & 0.49 ± 0.00 & 0.47 ± 0.00 & 0.54 ± 0.00 \\
     &  &  & Shuffle & 0.52 ± 0.01 & 0.47 ± 0.01 & 0.61 ± 0.02 \\
    \cline{1-7} 
    \multirow[t]{8}{*}{Variable} & \multirow[t]{4}{*}{Elitist} & \multirow[t]{2}{*}{1-Point} & Bit Flip & \textbf{0.87} ± 0.00 & \textbf{0.89} ± 0.00 & \textbf{0.90} ± 0.00 \\
     &  &  & Shuffle & 0.85 ± 0.00 & 0.78 ± 0.00 & 0.83 ± 0.01 \\
    \cline{3-7}
     &  & \multirow[t]{2}{*}{Uniform} & Bit Flip & 0.85 ± 0.00 & 0.78 ± 0.00 & 0.85 ± 0.00 \\
     &  &  & Shuffle & 0.84 ± 0.00 & 0.77 ± 0.00 & 0.82 ± 0.01 \\
    \cline{2-7} 
     & \multirow[t]{4}{*}{Tournament} & \multirow[t]{2}{*}{1-Point} & Bit Flip & \textbf{0.87} ± 0.00 & 0.86 ± 0.00 & 0.89 ± 0.00 \\
     &  &  & Shuffle & 0.85 ± 0.00 & 0.78 ± 0.00 & 0.84 ± 0.01 \\
    \cline{3-7}
     &  & \multirow[t]{2}{*}{Uniform} & Bit Flip & \textbf{0.87} ± 0.00 & 0.81 ± 0.00 & 0.88 ± 0.00 \\
     &  &  & Shuffle & 0.85 ± 0.00 & 0.78 ± 0.00 & 0.84 ± 0.01 \\
    \bottomrule
    \end{tabular}
\end{table}

\subsection{Hyperparameter Optimization}
The aim is to find solutions for the optimization problem in Equation~\eqref{eq:multiobjective} using the NSGA-II algorithm.
To find the best genetic operators for it,
we conducted hyperparameter optimization.
We use grid search to go through all configurations of genetic operators.
For each operator combination,
we evaluated the resulting Pareto front using the \emph{hypervolume indicator} (HV)~\cite{fonseca2006hypervolume}.
Due to the stochastic nature of GAs,
we conducted 10 trials for each combination.
We used a population size of 100 and the number of
generations was set to 200.
This was done on all given datasets.

\subsubsection{Hyperparameters}
There are several known methods we can choose from for
each genetic operator (\emph{initializer, selection, crossover, mutation})~\cite{goldberg1996genetic,introductionevolution}.
We only considered those that return binary vectors,
as our solutions are binary.
For all selection methods, we used two parents. Furthermore,
the bit flip mutation rate was set to 5\%.

\subsubsection{Metric}
To goal is to maximize the \emph{hypervolume indicator} (HV)~\cite{fonseca2006hypervolume}.
It measures the volume between the Pareto front and a reference point.
A higher hypervolume indicates more
coverage of the solution space and hence a better Pareto front.
We use the nadir point (1, 1) as the reference point. A value of 1 indicates
the theoretically best possible solution and a value of 0 indicates the worst.

\subsubsection{Results}
The results are displayed in Table~\ref{table:results_hyper}.
We display the mean and standard deviation of the hypervolume indicator for each dataset and genetic operator combination from the trials.
To answer \textbf{RQ1}, we observed best results
with variable initializer, elitist selection, 1-point crossover,
and bit flip mutation in our experiments.
Remarkably, the results confirm that the variable initializer suits our
problem better than the random initializer as it outperformed the latter in all datasets by a significant margin.
We also observed that elitist selection slightly outperformed
the binary tournament selection
method. Bigger differences can be observed in the crossover and mutation operators.
We note that all results were very consistent, as the maximum standard deviation was 0.01
and most often 0.00 when rounding to two decimals.

\subsection{Bias Mitigation and Classification Performance}
The aim of this experiment is to assess the performances of
various machine learning classifiers trained on datasets
pre-processed for fairness using our methodology.
Specifically, we aim to evaluate the effectiveness
in mitigating bias while maintaining classification accuracy.
For this, we also trained the models
on the unprocessed datasets, serving as a baseline.

\subsubsection{Train and Test Split}
For each dataset, we split the data into training and testing sets,
ensuring stratification based on sensitive attributes to preserve
representativeness across groups.
We used a 80-20 split for training and testing, respectively.

\subsubsection{Bias Mitigation}
After splitting, we either applied the multi-objective optimization approach or the single-objective optimization approach to mitigate
bias in the training data.
The test data was left completely unprocessed.

For both approaches, we experimented with different normalization factors
$\beta \in \{(1+\epsilon), \hat{\psi}(\mathcal{D})\}$.
The population size was set to 200 and the number of generations was set to 400.

For the multi-objective approach, we used the NSGA-II algorithm with the best genetic operators identified in the former experiment.
Because we get a Pareto front of solutions, we need to select one solution
for the evaluation.
We selected the solution from the Pareto front $\mathcal{PF}$ based on $\beta$ as follows:
\begin{equation*}
    \argmin_{\mathcal{D}_\text{fair} \in \mathcal{PF}} \quad \frac{\hat{\psi}_\text{SDP-max}(\mathcal{D}_\text{fair})}{\beta} + \mathcal{L}(\mathcal{D},\mathcal{D}_\text{fair}).
\end{equation*}

For the single objective, we used the genetic algorithm from Duong et al.~\cite{duong2023framework} with the same genetic operators
as the NSGA-II algorithm for comparison.
We used the solution that is returned when
solving the optimization problem in Equation~\eqref{eq:singleobjective}.
An alpha value of 0.5 was set to equally weigh the objectives.
However, we note that $\beta$ also influences the selection of the solution.

\subsubsection{Machine Learning Models}
We trained several machine learning classifiers implemented by the
\texttt{scikit-learn} library~\cite{scikit-learn},
including \emph{Logistic Regression} (LR),
\emph{Support Vector Machines} (SVM),
\emph{Random Forest} (RF), and
\emph{Artificial Neural Networks} (ANNs), on both the pre-processed fair data and the original, unprocessed data.

\subsubsection{Metrics}
Using the test set, we evaluated the models' predictions
on fairness and performance.
For fairness, we used the maximal statistical disparity
(see Example~\ref{ex:maxstatdisparity}).
We used this instead of the penalized version because
there is no need to penalize the test set for coverage.
We note that the test set automatically contains all groups due to the
stratification in the train-test split.
For the classifiers' performances, we report
the \emph{area under the receiver operating characteristic curve} (AUROC),
where higher values indicate better performance.

\subsubsection{Results}
\begin{table}[tb]
    \centering
    \caption{Fairness and relative size of the pre-processed training
    sets compared to the original training sets. Best results are in bold.}
    \label{table:results_train}
    \begin{tabular}{llrrr}
    \toprule
    Dataset & Approaches & $\hat{\psi}_\text{SDP-max}(\mathcal{D})$ & $\hat{\psi}_\text{SDP-max}(\mathcal{D}_\text{fair})$ & $|\mathcal{D}_\text{fair}|/|\mathcal{D}|$ \\
    \midrule
    \multirow[t]{4}{*}{Adult} & Single $\beta = \hat{\psi}(\mathcal{D})$ & $17\%$ & \textbf{5\%} & $54\%$ \\
     & Single $\beta = (1+\epsilon)$ & $17\%$ & $16\%$ & \textbf{99\%} \\
     & Multi $\beta = \hat{\psi}(\mathcal{D})$ & $17\%$ & $8\%$ & $54\%$ \\
     & Multi $\beta = (1+\epsilon)$ & $17\%$ & $17\%$ & \textbf{99\%} \\
    \cline{1-5}
    \multirow[t]{4}{*}{Bank} & Single $\beta = \hat{\psi}(\mathcal{D})$ & $25\%$ & \textbf{6\%} & $51\%$ \\
     & Single $\beta = (1+\epsilon)$ & $25\%$ & $25\%$ & \textbf{99\%} \\
     & Multi $\beta = \hat{\psi}(\mathcal{D})$ & $25\%$ & $11\%$ & $53\%$ \\
     & Multi $\beta = (1+\epsilon)$ & $25\%$ & $25\%$ & \textbf{99\%} \\
    \cline{1-5}
    \multirow[t]{4}{*}{Compas} & Single $\beta = \hat{\psi}(\mathcal{D})$ & $21\%$ & \textbf{1\%} & $54\%$ \\
     & Single $\beta = (1+\epsilon)$ & $21\%$ & $15\%$ & $94\%$ \\
     & Multi $\beta = \hat{\psi}(\mathcal{D})$ & $21\%$ & $3\%$ & $53\%$ \\
     & Multi $\beta = (1+\epsilon)$ & $21\%$ & $15\%$ & \textbf{96\%} \\
    \bottomrule
    \end{tabular}
\end{table}

\begin{figure*}[tb]
    \centering
    \subfloat{\includegraphics[width=\textwidth]{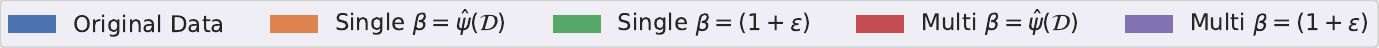}}\\
    \vspace{0.5em}
    \addtocounter{subfigure}{-1}%
    \subfloat[Adult (LR)]{\includegraphics[height=0.088\textheight]{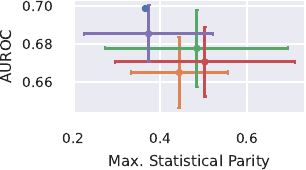}
    \label{fig:adult_lr}}%
    \subfloat[Adult (SVM)]{\includegraphics[height=0.088\textheight]{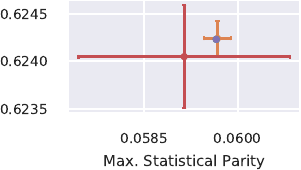}
    \label{fig:adult_svm}}%
    \subfloat[Adult (RF)]{\includegraphics[height=0.088\textheight]{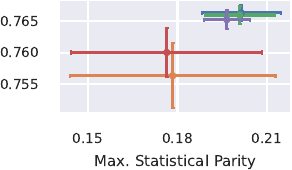}
    \label{fig:adult_rf}}
    \subfloat[Adult (ANN)]{\includegraphics[height=0.088\textheight]{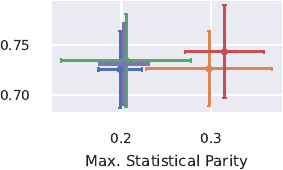}
    \label{fig:adult_ann}}

    \subfloat[Bank (LR)]{\includegraphics[height=0.089\textheight]{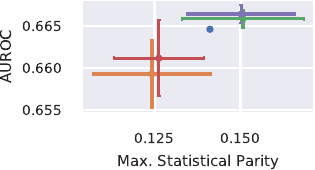}
    \label{fig:bank_lr}}%
    \subfloat[Bank (SVM)]{\includegraphics[height=0.089\textheight]{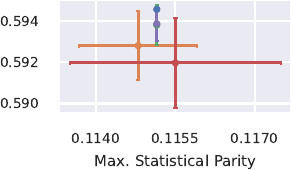}
    \label{fig:bank_svm}}%
    \subfloat[Bank (RF)]{\includegraphics[height=0.089\textheight]{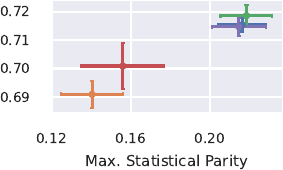}
    \label{fig:bank_rf}}
    \subfloat[Bank (ANN)]{\includegraphics[height=0.089\textheight]{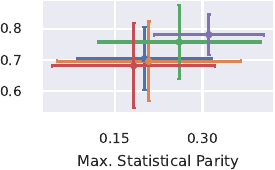}
    \label{fig:bank_ann}}

    \hfil
    \subfloat[Compas (LR)]{\includegraphics[height=0.09\textheight]{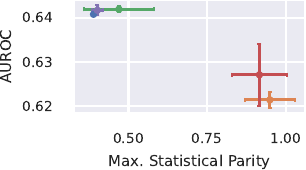}
    \label{fig:compas_lr}}%
    \subfloat[Compas (SVM)]{\includegraphics[height=0.09\textheight]{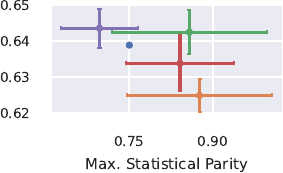}
    \label{fig:compas_svm}}%
    \subfloat[Compas (RF)]{\includegraphics[height=0.09\textheight]{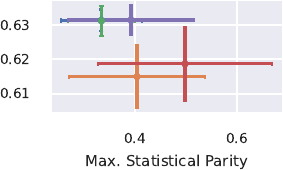}
    \label{fig:compas_rf}}
    \subfloat[Compas (ANN)]{\includegraphics[height=0.09\textheight]{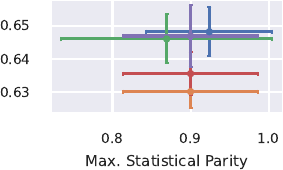}
    \label{fig:compas_ann}}
    \caption{Results on the test set using two approaches with varying $\beta$ parameters. x-axis and y-axis represent discrimination and
    performance metrics, respectively.}
    \label{fig:results_test}
\end{figure*}

Table~\ref{table:results_train} shows the maximal statistical disparity
values of the original and pre-processed training sets,
as well as the relative size of the pre-processed training sets
compared to the original training sets.
Notably, setting $\beta=\hat{\psi}(\mathcal{D})$ results in a
much lower discrimination than setting $\beta=(1+\epsilon)$,
but also in a higher data loss.
We observe only small differences between the single- and
multi-objective approaches in the discrimination values
when setting $\beta=\hat{\psi}(\mathcal{D})$.
The classifiers' results are displayed in Fig.~\ref{fig:results_test}.
We compare the classifiers' performances on the test set
using both the single- and multi-objective optimization approaches.
For each approach, we varied the normalization factor $\beta$ as described above.
We use error bars to display the mean and standard deviation of the AUROC and the maximal statistical disparity values from the 10 trials.

We do not observe a clear trend in the results regarding
improving or worsening fairness and performance.
This emphasizes that the pre-processed datasets can indeed
be used reliably as training data.
Only the experiment shown in Fig.~\ref{fig:compas_lr} stands out,
where the predictions on the pre-processed data are
significantly less fair than on the original data.

When comparing the classifiers, SVM seems to be least affected
by the pre-processed datasets, mostly apparent in Fig.~\ref{fig:adult_svm}
and Fig.~\ref{fig:bank_svm}.
Further, fairer classifiers tend to have lower performances,
indicating a trade-off between fairness and performance.

Contrasting the approaches with different $\beta$ parameters,
we do not see a clear winner.
Notable is the result of the single-objective approach with
$\beta = (1 + \epsilon)$ in Fig.~\ref{fig:compas_lr},
where its fairness did not worsen.
The experiment in Fig.~\ref{fig:bank_ann} shows an interesting
result, where less fair predictions achieve better performances.
This also indicates that a fairness-performance
trade-off is indeed available.

We can conclude that the pre-processed datasets can be used
reliably as training data, but an improvement in fairness
can only be guaranteed in the training set
and not necessarily in the test set. 
Evidence of a fairness-performance trade-off was found in the results.
This explains why fairness was rarely improved in the experiments
and why the classifiers' performances were not negatively affected.
The protected attributes seem to correlate with the target variable,
making it difficult to remove the discrimination without
changing the data significantly.

\section{Discussion}
In this section, we reflect on key aspects of our approach, particularly focusing on data quality and the trade-off between fairness and data loss.

\subsection{Data Quality}
We extend the work of Duong et al.~\cite{duong2023framework}
by introducing new constraints and objectives that the
resulting dataset must fulfill.
To our knowledge, we are the first to specifically aim for data quality
while pre-processing datasets for fairness.
Some techniques in the literature~\cite{fairlearn,aif360}
violate several data integrity constraints and transform the dataset
in a way that makes it unusable.
For example, \texttt{CorrelationRemover} in
\texttt{Fairlearn}~\cite{fairlearn}
projects discrete features into continuous features.
Categorical features such as the label are also affected by it,
making it impossible to train classifiers for comparative purposes.
The pre-processed datasets from our work
do not come with these issues,
making them more useful in practice.

\subsection{Trade-off}
We proposed a multi-objective optimization problem
where the objectives are fairness and data loss.
Solving this problem results in a set of Pareto-optimal solutions
where each solution is a fair subset.
The user can then choose the most suitable subset for their application
using the weighted sum of the objectives.
Here, we proposed a parameter $\beta$ that weights the fairness objective.
The choice of $\beta$ has to be made by the user and depends on
the size of the dataset and the importance of fairness in the application.
Our suggested values for $\beta$ serve as initial guidance, allowing users to further adjust based on their needs.

\section{Conclusion}
In this paper, we developed a data pre-processing technique
that aims to remove discriminating data points for fairness
while maintaining data quality.
We introduced two additional criteria, \emph{coverage} and \emph{data loss}, to enhance the trustworthiness of the resulting dataset.
Using our methods, the fairness of the dataset can be improved
without compromising the quality of the data.
By evaluating our methodology on machine learning models
using three real-world datasets, the results show that the models'
fairness and performances were not
affected significantly by the data removal process
compared to models trained on the original datasets.
This indicates that the pre-processed datasets are reliable and can be used for further analysis.


%
%
%
\bibliographystyle{splncs04}
\bibliography{references}
\end{document}